# Interpretable Drug Synergy Prediction with Graph Neural Networks for Human-AI Collaboration in Healthcare

Zehao Dong[1], Heming Zhang[1], Yixin Chen[1], Fuhai Li [2,3]

[1]Computer Science, Washington University in St. Louis, St. Louis, MO, USA.

[2]Institute for Informatics (I2), Washington University School of Medicine, Washington University in St. Louis, St. Louis, MO, USA.

[3]Department of Pediatrics, Washington University School of Medicine, Washington University in St. Louis, St. Louis, MO, USA.

zehao.dong@wustl.edu; hemingzhang@wustl.edu; chen@cse.wustl.edu; Fuhai.Li@wustl.edu

## Abstract

We investigate molecular mechanisms of resistant or sensitive response of cancer drug combination therapies in an inductive and interpretable manner. Though deep learning algorithms are widely used in the drug synergy prediction problem, it is still an open problem to formulate the prediction model with biological meaning to investigate the mysterious mechanisms of synergy (MoS) for the human-AI collaboration in healthcare systems. To address the challenges, we propose a deep graph neural network, IDSP (Interpretable Deep Signaling Pathways), to incorporate the gene-gene as well as gene-drug regulatory relationships in synergic drug combination predictions. IDSP automatically learns weights of edges based on the gene and drug node relations, i.e., signaling interactions, by a multi-layer perceptron (MLP) and aggregates information in an inductive manner. The proposed architecture generates interpretable drug synergy prediction by detecting important signaling interactions, and can be implemented when the underlying molecular mechanism encounters unseen genes or signaling pathways. We test IDWSP on signaling networks formulated by genes from 46 core cancer signaling pathways and drug combinations from NCI ALMANAC drug combination screening data. The experimental results demonstrated that 1) IDSP can learn from the underlying molecular mechanism to make prediction without additional drug chemical information while achieving highly comparable performance with current state-of-art methods; 2) IDSP show superior generality and flexibility to implement the synergy prediction task on both transductive tasks and inductive tasks. 3) IDSP can generate interpretable results by detecting different salient signaling patterns (i.e. MoS) for different cell lines.

## 1.Introduction

In recent years, there have been important problems arising from the need for institutions to develop interpretable human-AI collaboration systems to facilitate synergistic and effective drug combination discovery for cancer treatment. Most of existing deep-learning based drug synergy models concatenate a large set of omics data and chemical structures data for the prediction. These models are hard to be interpreted, thus, they fail to incorporate human expertise in the diagnosis. It has been shown that cancer is driven by genetic and epigenetic alterations, many of which can be mapped into signaling pathways that controls the survival and migration/invasion of cancer cells. As one reported drug pair analysis indicates,

89% of tumor samples had at least one driver alteration in one of 10 cancer-related signaling pathways that is responsible for the tumor development, while 57% and 30% had one and multiple potentially druggable targets respectively. Thus, inhibited signaling targets analysis shows great potential of facilitating the synergistic drug combinations discovery. However, the mechanisms of synergy (MoS) is heterogeneous for different drug combinations in different diseases, while the reported signaling pathways for cancer therapy are rather limited and there are only small sets of feasible drugs, thus it is challenging to systematically model the drug combination synergy from the system biology perspective to support inductive prediction tasks with unseen gene-gene and gene-drug interactions. Hence, AI architectures that incorporate the underlying synergistic and antagonistic molecular mechanism show great potential of unleashing the power of interpretable AI system in medical analysis to provide the comprehensive and deep insight into the MoS of drug combinations for better cancer patients outcome.

The past decade has witnessed the great success of deep learning/ machine learning methods on applications in the domain of drug combination discovery for cancer therapies. For instance, several proposed machine learning models build ensembles tree to predict the synergy scores (TreeCombo, Random Forest). While various deep neural networks are developed to learn the node embeddings based on chemical and gene expression profiles in specific cell lines, and then the learnt drug embeddings are used to predict the drug pair synergy (**DeepSynergy**, MatchMaker). Though these approaches have achieved expressive results in the drug combination prediction task, they still share a major limitation that they fail to model the gene-gene interactions on the gene regulatory signaling networks nor the gene-drug interactions, which makes it difficult to interpret the prediction in a biological meaningful order.

To incorporate molecular interactions in drug synergy analysis, graphs are used to represent the complexity in these systems, and graph neural networks (GNNs) have been receiving more and more attentions in recent years to effectively extract useful information from the graph-structured system, such as physical system, protein-protein interaction networks, gene-gene interaction networks (signaling pathways), etc. For instance, Zitnik et al. modeled the side effects of the drug combinations by predicting whether a specific type of link exits in the incomplete networks of drugs and proteins, and the proposed model successfully improves the prediction accuracy and achieves the state-of-art performance. However, the model fails to detect important drug-protein nor protein-protein interactions that have biological meaning and can explain the generated outcome, thus it prevents the collaboration between the proposed AI algorithm and experts to address the human needs in healthcare. In fact, Pan et al,. has shown that the dug combination of Venetoclax and Idasanutlin can generate antileukemic efficacy in the treatment of acute myeloid leukemia by inhibiting antiapoptotic Bcl-2 family proteins and activating the p53 pathway at same time. In practice, unfortunately, it is impossible to manually identify all critical factors in diagnosis due to the cost and vast number of drug pairs. Hence, interpretable deep models that explain underlying molecular mechanism are in an urgent need for AI-assisted healthcare systems.

Besides the loss of interpretability, another major challenge in the drug synergy prediction problem is to develop the inductive model which is feasible and effective when molecules (i.e. proteins, genes, enzymes, etc) not encountered in the training data are allowed to appear during the testing process. Though various databases have been formulated, they only contain small proportional molecular connections that are experimentally verified As such, the proposed AI algorithm has to be able to deal with unseen molecules when applied in real-world healthcare systems.

To design an interpretable model that supports inductive drug synergy prediction in real-world cancer therapy discovery, we propose a novel inductive graph neural network, IDWSP (Inductive Deep Weighted Signaling Pathways), to address aforementioned challenges and provide interpretable synergic drug combination prediction. In IDWSP, the edge importance is inductively learnt by a multiple layer perceptron (MLP) to measure the importance of signaling pathways. Then the drug node embeddings are generated through neighborhood aggregation with weighted adjacency matrix formulated by the learnt edge importance, and the generated node embeddings are used to predict the synergic score in a symmetric way. We validate the proposed model on drug combinations from NCI ALMANAC, and the proposed architecture shows highly comparable results with state-of-art approaches on transductive prediction tasks and achieves competitive result when applied in inductive learning tasks with unseen gene-gene interactions and gene-drug interactions.

## 2. Related Works

**Graph neural networks**: Graph neural networks (GNNs) have successfully applied deep neural networks to learning tasks over structured data such as graphs. GNNs mainly are generalized from convolution neural networks (CNNs) by designing graph convolution and graph pooling to unleash the representation learning ability. The graph convolution aggregates information from spectral or spatial perspective to generate node representations that encode both the graph topology and the node feature. The graph pooling module extract the graph representation based on generated node representations for graph-level tasks like graph classification. GNNs have been proven to be effective in both node representation learning and graph representation learning, and have achieved current state-of-art performance in various graph tasks, such as node classification, link prediction and graph classification.

**Machine Learning in drug synergy prediction:** The drug synergy analysis is difficult to implement manually, since chemical experiments are costly and are only feasible for small set of drugs. In addition, due the vast number of drug combinations and limited experimental screening data, it is practically impossible to evaluate all drug combinations in a biologically meaningful manner. Hence, computational models apply machine learning / deep learning algorithms to predict the synergic score of drug pairs. To improve the generality and flexibility over the feature preprocessing based models that operate on large compound feature library including chemical drug structures, omics data, etc, several deep learning models are proposed to make prediction based on the drug chemical structure and cell line gene expression. And have achieved current state-of-art performance. DeepSynergy concatenates the chemical descriptors of tested drug pair and cell line gene expression profile as input, and feeds the concatenated feature into a MLP to predict the corresponding synergy score. Inspired by DeepSynergy, Matchmaker takes the same input, but separately utilizes two drug specific networks (i.e. MLPs) learns the embedding of each drug based on the which concatenated feature of corresponding drug chemical structure and cell line gene expression. However, these models are unable to model the underlying gene interactions in signaling pathways, thus they fail to interpret the underlying molecular interactions for human-AI collaboration. To address this challenge, DeepSignalingSynergy designs a deep MLP model that incorporates gene-gene interactions and drug-gene interactions. DeepSignalingSynergy takes 1648 genes features (i.e. gene expression, copy number, target gene of drug 1, target gene of drug 2) as input, thus it predict the drug combination synergy score without drug chemical features. However, DeepSignalingSynergy is inherently transductive and is not feasible to drug pairs that have target genes not included in the selected 1648 genes.

## 3. Methodology

In this section, we outline the architecture of the proposed Interpretable Deep Signaling Pathways (IDSP), and show how it detects the biological meaningful gene regulatory relationship for synergic drug combinations. IDSP contains two main phases: In phase 1) we utilize graph convolution to enclose rooted subtrees around the drug pairs in the input graph and to model gene regulatory relations and interactions between drug and gene. The overall structure is illustrated in the Figure 2. More specifically, we utilize an attention-based weight-learning machine to learn the connection strengths (i.e. weights) between molecules in the input signaling network at the beginning of the learning process, and the learnt strengths play the role as weights of the neighborhood aggregation in the first graph convolution layer. In phase 2), the extracted node embeddings of two drug nodes are passed into a synergy predictor to map the encoded subtrees around the drug pairs into corresponding synergy scores.

## 3.1 Problem formulation

In this work, we study the inductive settings where molecular (i.e. genes and drugs) networks are used for the synergy prediction task in the drug discovery, while not all the genes are available during the training process. This problem setting well characterize many real-world scenarios in the Human-AI collaboration system, since it's practically impossible to involve all signaling pathways when the AI algorithm is trained. The existence of missing genes could be extremely vital for the high-quality model performance. For instance, as Figure 1 indicates, gene 6 has the highest centrality yet missed in the training phase, then the AI model could be forced to focus on molecular relations that play the less important role in the purpose of the synergy prediction. To address this challenge, the prediction model is required to be able to inductively encode the importance of substructures in the information propagation process.

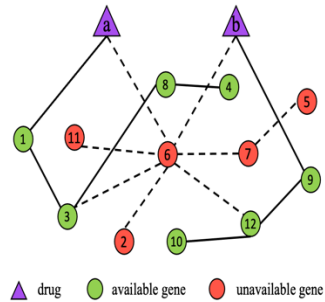

Figure 1: Illustration of the inductive pathway-based synergy prediction

The inductive synergy prediction problem takes as input a set of graphs $(G_1, G_2, ..., G_N)$ consist of genes and drug pairs, and the tasks is to predict their corresponding synergy scores. The dataset is separated into a training set and a testing set in each run of cross validation sets, and a specific proportion (10%) of genes in $(G_1, G_2, ..., G_N)$ are completely unobserved during training.

## 3.1 Attention-based weight-learning machine

We will start by introducing the attention-based weight-learning machine that learns the influence between molecules in an inductive manner. Since the gene order and the number of gene in the input molecular network are inconsistent, the proposed weight-learning machine should be invariant to the permutation of the input node order, and be able to deal with input graphs with different sizes as well.

Thanks to the universal approximation theorem (Hornik et al., 1989; Hornik, 1991) and the graph attention network that quantifies the weights of in neighborhood aggregation process to propagate information in graphs, we propose a weight-learning mechanism with following form:

$$e_{i,j} = relu(\text{MLP}(concat(h_i, h_i))) \quad (1)$$

$$\begin{cases} h_k = P_1 x_k & \text{if node k is a drug node} \\ h_k = P_2 x_k & \text{if node k is a gene node} \end{cases}$$

$$v_{i,j} = \frac{e_{i,j}}{\max\{e_{i,j} | i \in \mathcal{N}_r(j)\}} \quad (2)$$

Where $(x_1, x_2, ..., x_n)$ is the set of node features in the input network, and $P_1, P_2$ are trainable parameter matrices that provides sufficient expressive power to reflect the complexity of the molecular network. More specifically, we separately model the effects of different types of nodes through different parameter matrices $P_1, P_2$ to encode their different roles in the signaling pathways. Compared to other nodes, nodes incident to edge (i,j) play more critical roles in the edge importance analysis, thus we take the concatenation of drug node embeddings as the input to a multiple layer perceptron.

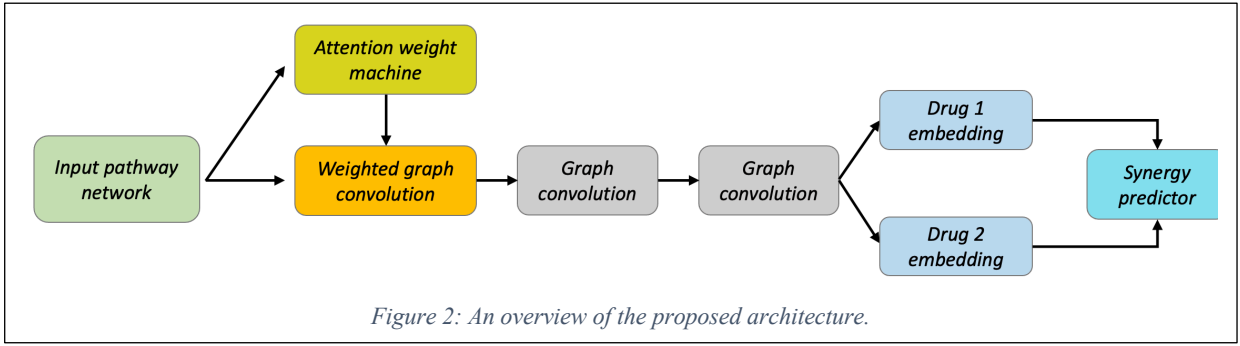

Figure 2: An overview of the proposed architecture.

### 3.2 Relational graph convolution layers

In order to learn the rich patterns introduced by different edge types (gene-gene interaction and gene-drug interaction) and node types (gene nodes and drug nodes), we combine the relational graph convolutional operator (R-GCN) and message passing neural network (MPNN) as the proposed GNN message passing layer, which takes the form as following:

$$z_j^{(l+1)} = relu(\text{concat}\left(M_0^{(l)} z_j^{(l)}, \sum_{r \in \Phi} \sum_{i \in \mathcal{N}_r(j)} \frac{1}{|\mathcal{N}_r(j)|} w_{i,j}^{[l]} h_i^{(l)}\right)) \quad (2)$$

$$\begin{cases} h_k^{(l)} = M_1^{(l)} z_k^{(l)} & \text{if node k is a drug node} \\ h_k^{(l)} = M_2^{(l)} z_k^{(l)} & \text{if node k is a gene node} \end{cases} \quad (3)$$

Where $z_j^{(l+1)}$ denotes the features of node j in layer $(l+1)$; $\Phi = \{$gene-gene edge, drug-gene edge$\}$ is the set of edge types; $\mathcal{N}_r(j)$ denotes the set of nodes that are linked with node $j$ through type $r$ edge. while $w_{i,j}^{[r]}$ is the learnt edge weight (importance) for a specific type of edge between node $i$ and node $j$. For better interpretability, we simply model the connection between molecules in the input graph, but not that between localized graph sub-structures around them. That is, we only utilize the learnt weights in the neighborhood aggregation process in the first graph convolution layer (i.e. $w_{i,j}^{[0]} = v_{i,j}$), and all the neighboring nodes in other graph convolution layers are equally treated (i.e. $w_{i,j}^{[l]} = 1, l > 0$).

In the equation (2), $M_0^{(l)} z_j^{(l)}$ and $\sum_{r \in \Phi} \sum_{i \in \mathcal{N}_r(j)} \frac{1}{|\mathcal{N}_r(j)|} w_{i,j}^{[l]} h_i^{(l)}$ respectively represent the embedded information of the central molecule (node) and the information passed in through the signaling pathways, and we concatenate them to distinguish the central node from the neighboring nodes. Since information aggregation through different interactions are processed with different weight matrix ($M_1^{(l)}$ and $M_2^{(l)}$), such

message passing layer provides expressive power to reflect the massive interaction patterns of the enriched physical interactions in the input graph.

### 3.3 Symmetric synergy predictor

To explore the depth and the breadth of the receptive fields of nodes, we stack multiple (L) proposed message passing layers to generate the node embeddings. Then the generated drug node embeddings in the last layer ($x_{drug\ 1}^{(L-1)}, x_{drug\ 2}^{(L-2)}$) are used as the graph representation and are feed into a symmetric synergy decoder to predict the synergy scores.

Ideally, we expect the decoder model to be invariant to the order of the drug node embeddings, i.e. the drug pair (drug A, drug B) should have same synergic score as the drug pair (drug B, drug A). Here we utilize a symmetric decoder similar to the factorization decoder in decagon[23] as following:

$$Score = (x_{drug\ 1}^{(K)})^T D^T D (x_{drug\ 2}^{(K)}) \quad (5)$$

In the proposed symmetric decoder, the trainable parameter matrix $D$ models relations between different channels in the generated drug node embeddings, thus it provides better expressive power compared to the simple inner product of two drug node embeddings $(x_{drug\ 1}^{(K)})^T x_{drug\ 2}^{(K)}$. Furthermore, the symmetric synergy predictor is performed through matrix multiplication and supports end-to-end learning, thus it is efficient and effective.

## 4. Experiments

We evaluate IDSP by comparing the performance against a wide variety of powerful baselines in the transductive settings. To show the strong inductivity of the proposed architecture, we compare it's performance under the inductive setting and that under the transductive setting. Besides above attempts to measure the expressive power of proposed model in the different synergy prediction tasks, we also show implement exploratory analysis to show that it could be beneficial to involve the molecule interactions in the synergy prediction.

### 4.1 Dataset

We formulate the synergic drug prediction task as an inductive regression problem on input graphs of two node types: the gene nodes and the drug nodes, and Figure 1 provides an example of the input graph to IDWSP. These gene nodes are collected from 46 famous signaling pathways/networks (45 "signaling pathways" + cell cycle)[8] in KEGG (Kyoto Encyclopedia of Genes and Genomes)[10] database, while the NCI Almanac dataset contains the combo-scores[1] of drug combinations of 104 FDA approved drugs against the tumor growth of NCI60 human tumor cell lines, based on which the synergic score of drug combinations can be omputed as the average combo-score with different doses on a given tumor cell line.

**Gene-gene interaction**: The gene-gene interaction network encodes the physical signaling interactions from documented medical experiments, and such interaction relations are obtained from the KEGG dataset, including following 46 signaling pathways: : MAPK, FoxO, TGF-beta, T cell receptor, Adipocytokine, ErbB, Sphingolipid, VEGF, B cell receptor, Oxytocin, Ras, Phospholipase D, Apelin, Fc epsilon RI, Glucagon, Rap1, p53, Hippo, TNF, Relaxin, Calcium, mTOR, Toll-like receptor, Neurotrophin, AGE-RAGE, cGMP-PKG, PI3K-Akt, NOD-like receptor, Insulin, Cell cycle, cAMP, AMPK, RIG-I-like receptor, GnRH, Chemokine, Wnt, C-type lectin receptor, Estrogen, NF-kappa B, Notch, JAK-STAT, Prolactin,

HIF-1, Hedgehog, IL-17, Thyroid hormone; And these signaling pathways provides interaction relations between 1634 genes.

**Drug-gene interaction:** We obtain the drug-gene interaction from DrugBank[9] database (version 5.1.5, released 2020-01-03). There are 21 drugs in the NCI Almanac dataset with documented targets on the above 1634 signaling pathways genes, thus this research is built on drug combinations of these 21 FDA approved drugs: Celecoxib, Gefitinib, Quinacrine hydrochloride, Tretinoin, Cladribine, Imatinib mesylate, Romidepsin, Vinblastine sulfate (hydrate), Dasatinib, Lenalidomide, Sirolimus, Vorinostat, Docetaxel, Mitotane, Sorafenib tosylate, Thalidomide, Everolimus, Nilotinib, Tamoxifen citrate, Paclitaxel, Fulvestrant.

### 4.2. Exploratory analysis and data-driven motivation

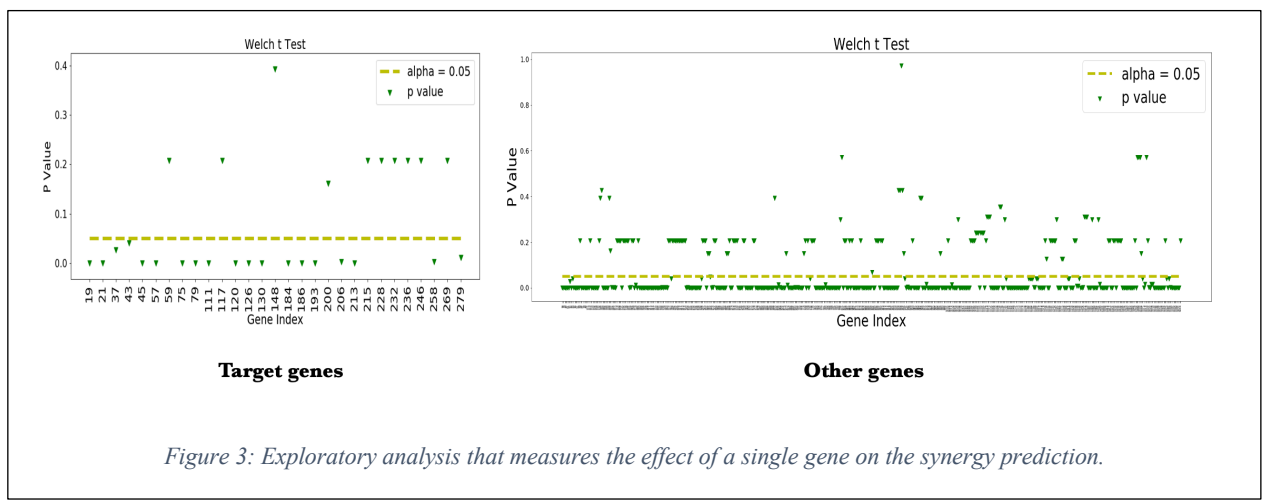

*Figure 3: Exploratory analysis that measures the effect of a single gene on the synergy prediction.*

Though the preclinical experimental efforts in the synergism discovery of cancer drug combination have proven that some targeting genes/ pathways are responsible for the drug resistance reduction, there is no computational analysis, to our knowledge, implemented to justify that modeling the gene integration and drug-target mechanism is beneficial for the synergy analysis. In order to do so, we need to answer following two questions:

1. Whether including specific type of gene-gene relations or drug-gene connection can cause significant difference on the synergy score?
2. What is the scope for the simplified model on a small set of signaling pathways compared to more complicated deep learning model using both drug chemical structure data and cell line profile data?

Here we implement t Welch t test to address the first concern, and the second concern is addressed by comparing IDWSP with other baselines.

**Gene-gene interaction**: We firstly focus on the genes from the 1-hop neighborhood of the input drug pairs (i.e. target gene). In the 46 signaling pathways, there are in total 29 genes as the target gene of at least one drug. We separate are all available drug pairs into two sets according to whether a specific gene is in the target gene set of the drug pairs, and record the corresponding synergy score. We consider a gene significantly has no impact on the synergy prediction if the t-test is at the significance level 5% can not distinguish the two sets of synergy scores, otherwise we think the impact of that gene is significant. As

Figure 3 shows, 10 of 29 genes has a t value larger than 0.05, thus the corresponding drug-gene interactions will significantly influence the synergy score of the drug combination.

**Drug-gene interaction:** Similar to previous analysis, we also study the effect of the gene-gene interactions. We also test whether non-target genes in the two hop neighborhood show significant effect on the synergy prediction. There are in total non-target 546 genes that satisfy the above condition, 149 of which have a t value larger than 0.05. Such observation indicates that only a proportion of the gene-gen interactions in the two-hop neighborhood of drug pairs have significant influence.

Above analysis can indicate that it would be beneficial to obtain the molecule structure into the synergy prediction task. However, there is an extremely large variety of molecule connections (i.e. gene-gene and gene-drug interactions) due to the large amount of genes and drug. Here comes the question that how these connections interact with each other and what's their effects on the synergy score.

### 4.3 Experimental Results

**Performance in transductive settings**: In the transductive setting, all genes are available in the training phase. We report the baseline results provided in MatchMarker, and Table 1 lists the experimental results. (1) Compared to DeepSynergy, IDSP utilizes no drug chemical features, yet have achieved highly comparable results with DeepSynergy. (2) DeepSignalingSynergy takes the same input as IDSP, yet the prediction model assumes that both the number and the order of genes in the input molecule network are consistent, thus it could be hard to generalize DeepSignalingSynergy to settings with unseen gene or settings with different number of genes. Furthermore, DeepSignalingSynergy doesn't provide any interpretable results that illustrate the molecule interaction mechanism. As such, it could be hard to implement DeepSignalingSynergy in the real-world AI-assisted healthcare. (3) We also compare IDSP with many baselines (i.e. TreeCombo, RandomForest) based large compound feature library including chemical drug structures, omics data, etc. As Table 1 indicates, IDSP can achieve highly competitive results in thee transductive settings, while using less information. (4) Both ElasicNet and the Baseline take the sane input as DeepSynergy. IDSP improve the Pearson correlation over these two baselines by a margin of around 0.17.

| Model | Pearson Correlation | Interpretability | Inductivity | Chemical drug input feature |
|---|---|---|---|---|
| **IDSP** | 0.64 | True | True | False |
| DeepSynergy | 0.68 | False | False | True |
| DeepSignalingSynergy | 0.63 | False | False | False |
| TreeCombo | 0.62 | True | True | True |
| RandomForest | 0.60 | True | True | True |
| ElasticNet | 0.44 | False | False | True |
| MLP | 0.43 | False | False | True |

Observations (1), (2), (3), and (4) indicates that modeling the molecule connections in the signaling pathways can unleash the full potential of AI in investigating mechanism of synergy (MoS). With additional structural information in the molecule network, IDSP can achieve highly comparable results while using less features.

**Performance in inductive settings**: Here we compare the performance of IDSP in inductive settings and in transductive settings. In the inductive setting, we set 10% of genes in the selected 46 signaling pathways unseen, i.e. around 5 signaling pathways of 46 selected signaling pathways are set as unavailable during the training phase. As Table 2 shows, IDSP can achieve comparable performance even if 10% of genes are unavailable, thus the proposed architecture successfully capture the underlying mechanism helpful for purpose of the synergy prediction.

| **Setting** | Mean square error | Pearson |
|---|---|---|
| Inductive | 45.56 | 0.59 |
| Transductive | 40.85 | 0.64 |

*Table 2: Model performance in the inductive setting and in the transductive setting*

### 4.4 Visualization

In order to evaluate the interpretability of IDSP, we visualize the detected salient molecular interactions (i.e. gene-gene interactions, gene-drug interactions) in the signaling network for drug pairs targeting on different cell lines. (1) Since we take 3 graph convolution layers in IDSP and predict the synergy score based two generated drug node embeddings, the prediction process is only dependent on the gene node from which the shortest path to any of drug node pairs is shorter than 3. Thus, it's reasonable to neglect rest gene nodes. (2) For better interpretability, we only model the edges importance in the first graph convolution layer, as they straightforward represent the connection strength between pair of nodes in the input network. On the other hand, since node representations in deeper graph convolution layers (i.e. layer $\geq 2$) encodes rooted subtrees instead of nodes in the network, we assume the edge importance in these graph convolution layers are equal (i.e. unweighted edges), thus the corresponding graph convolution layer only propagate information based on the network topology. As such, we only visualize the edges whose edge importance are larger than 0.5 in the first layer to provide interpretable MoS (Mechanism of Synergy) for comparison. (3) In graphs, nodes with high cardinality empirically play more critical roles in the graph representation learning. Thus, we could compare high-centrality nodes in different networks/graphs to distinguish different synergy mechanism.

To compare the synergy mechanism for different cell lines, we randomly select two cell lines: T-47D and NCI-H460, and then randomly choose 10 drug pairs on each cell lines. Here we take the first three drug pairs of each cell line and present the visualization results. In these visualized graphs, drugs are visualized as red nodes, while green nodes represent genes.

- T-47D:

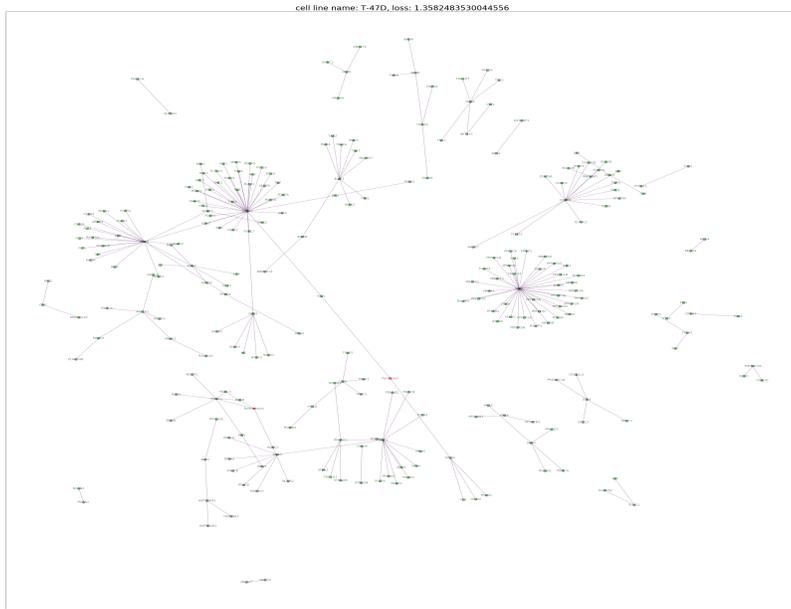

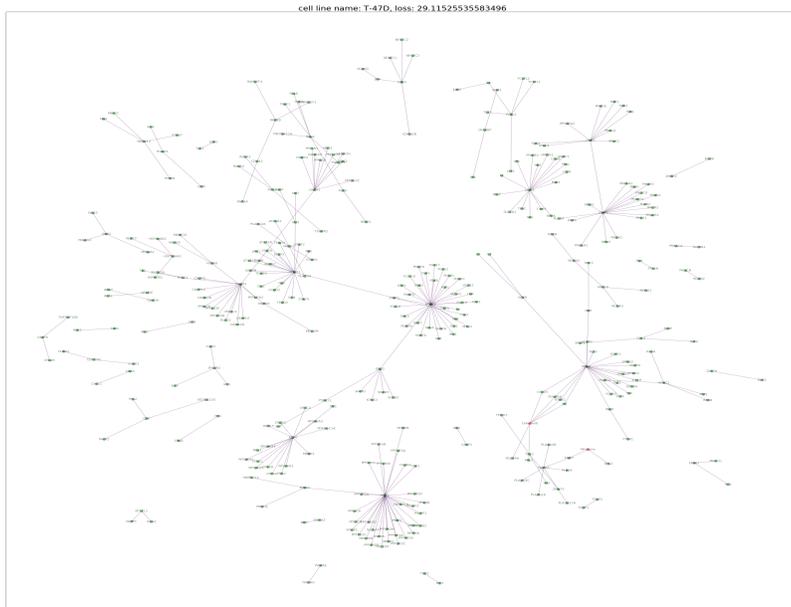

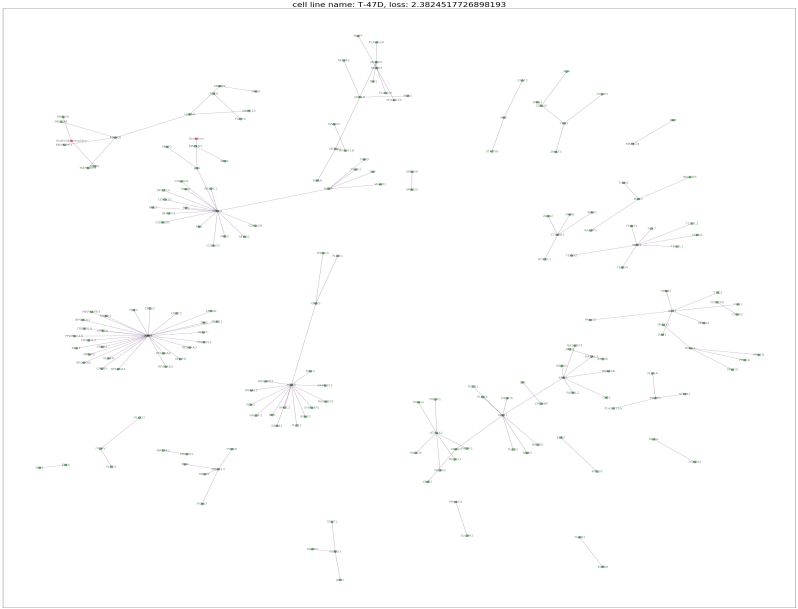

- NCI-H460:

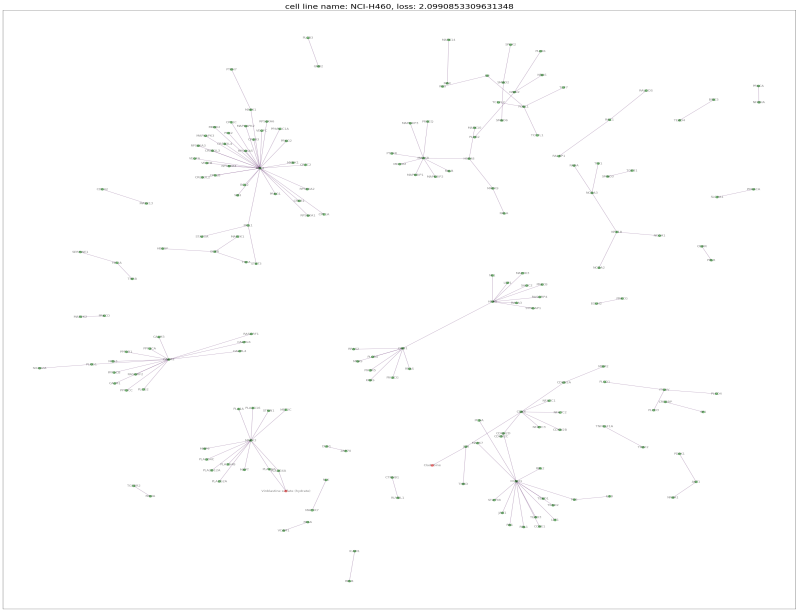

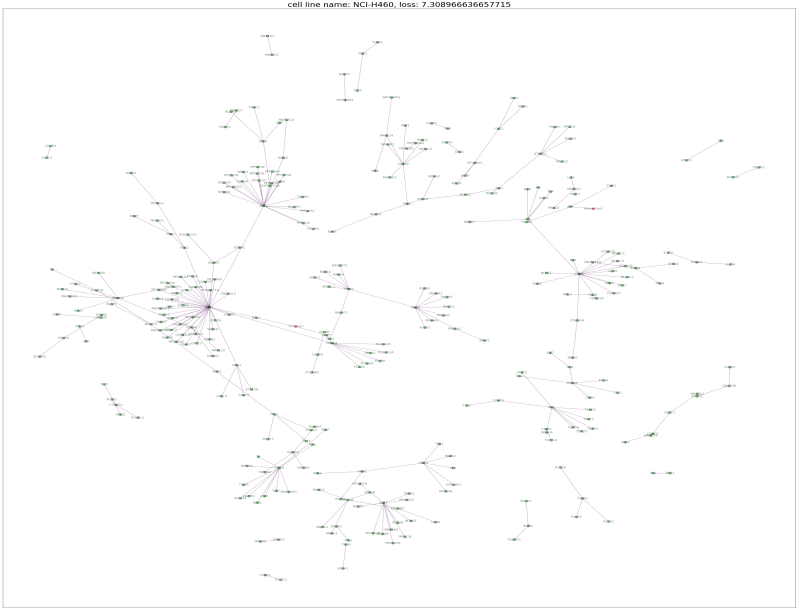

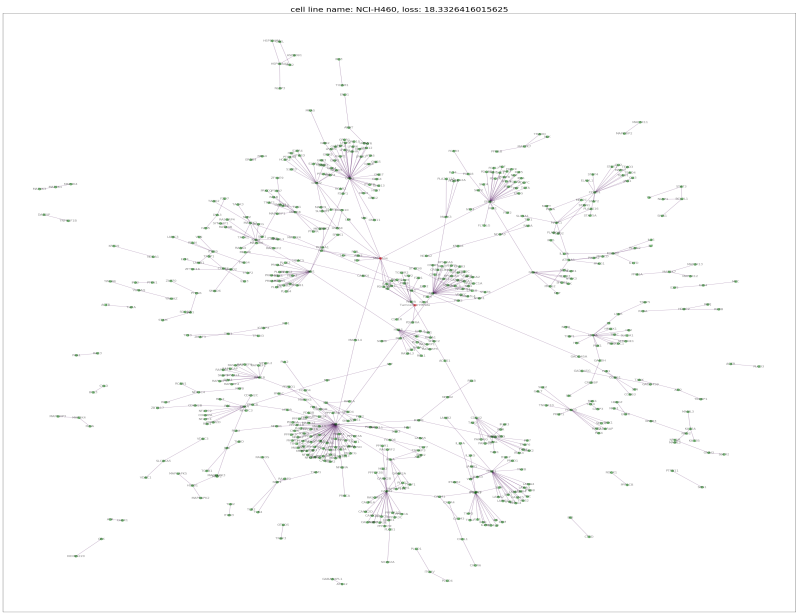

As these visualization suggest, cell line T-47D always detects high-centrality genes: AKT1, ATF4, GNB1, HRAS, EPHA2, GRB2, HRAS, as well as their corresponding localized signaling patterns (i.e. rooted subtrees of depth 1 or 2); while cell line NCI-H460 always detects: AKT2, ATF4 MARK3, CDK4, GNB1, HRAS. Though other genes and corresponding patterns also have relatively high centrality and may also help to predict the synergy scores, they are not as representative as selected genes. Then we have following observations:

1. $\frac{4}{7}$ detected important genes (AKT1, EPHA2, GRB2, HRAS) of cell line T-47D do not appear in that of cell line NCI-H460; While $\frac{1}{2}$ detected important genes (AKT2, MARK3, CDK4) of cell line NCI-H460 do not play critical mechanic roles for cell line T-47D.

2. Though ATF4 is detected by both T-47D and NCI-H460, the co-appearance patterns are different. For T-47D, ATF4 shows close relation to AKT1, yet it always appears with MARK3 for NCI-H460.

## 5. Conclusion

In this paper, we have proposed an interpretable GNN architecture called IDSP (Interpretable Deep Signaling Pathways) to predict the synergy score of drug combinations and to investigate the underlying mechanism of synergy (MoS) by detecting salient molecular interactions and patterns. Without any drug chemical information, IDSP not only achieves highly comparable results with current state-of-art baseline approaches that utilize drug chemical features, but also interpret the prediction from the perspective of signaling pathways. Massive experimental results show that incorporating cancer cell lines in the predictive model has great potential of developing interpretable AI systems for anticancer drug combination discovery in the Human-AI collaborative healthcare.

## References

1. E. Choi, C. Xiao, W. Stewart, and J. Sun, "Mime: Multilevel medical embedding of electronic health records for predictive healthcare," in *Advances in Neural Information Processing Systems*, 2018, pp. 4548–4558.

2. A. Fout, J. Byrd, B. Shariat, and A. Ben-Hur, "Protein interface prediction using graph convolutional networks," in *NIPS 2017*, 2017, pp. 6530–6539.

3. A. Sanchez-Gonzalez, N. Heess, J. T. Springenberg, J. Merel, M. Riedmiller, R. Hadsell, and P. Battaglia, "Graph networks as learnable physics engines for inference and control," *arXiv preprint arXiv:1806.01242*, 2018.

4. P. Battaglia, R. Pascanu, M. Lai, D. J. Rezende *et al.*, "Interaction networks for learning about objects, relations and physics," in *NIPS 2016*, 2016, pp. 4502–4510.

5. J. D. Janizek, S. Celik, and S.-I. Lee, "Explainable machine learning prediction of synergistic drug combinations for precision cancer medicine," bioRxiv, p. 331769, 2018.

6. T. Chen and C. Guestrin, "Xgboost: A scalable tree boosting system," in Proceedings of the 22nd acm sigkdd international conference on knowledge discovery and data mining. ACM, 2016, pp. 785–794.

7. Sanchez-Vega F, Mina M, Armenia J, et al. Oncogenic Signaling Pathways in The Cancer Genome Atlas. *Cell*. 2018;173(2):321-337.e10. doi:https://doi.org/10.1016/j.cell.2018.03.035

8. Feng J, Zhang H, Li F. Investigate the relevance of major signaling pathways in cancer survival using a biologically meaningful deep learning model. *bioRxiv*. January 2020:2020.04.13.039487. doi:10.1101/2020.04.13.039487

9. Wishart DS, Feunang YD, Guo AC, et al. DrugBank 5.0: A major update to the DrugBank database for 2018. *Nucleic Acids Res*. 2018;46(D1):D1074-D1082. doi:10.1093/nar/gkx1037

10. Ogata H, Goto S, Sato K, Fujibuchi W, Bono H, Kanehisa M. KEGG: Kyoto encyclopedia of genes and genomes. *Nucleic Acids Res*. 1999:28. doi:10.1093/nar/27.1.29

11. Zhang T, Zhang L, Payne P, Li F. Synergistic Drug Combination Prediction by Integrating Multi-omics Data in Deep Learning Models. *arXiv Prepr arXiv181107054*. 2018.

12. Kurt Hornik, Maxwell Stinchcombe, and Halbert White. Multilayer feedforward networks are universal approximators. *Neural networks*, 2(5):359–366, 1989.

13. Kurt Hornik. Approximation capabilities of multilayer feedforward networks. *Neural networks*, 4(2): 251–257, 1991.

14. Michael Schlichtkrull, Thomas N Kipf, Peter Bloem, Rianne Van Den Berg, Ivan Titov, and Max Welling. Modeling relational data with graph convolutional networks. In European Semantic Web Conference, pp. 593–607. Springer, 2018.

15. Stark C. BioGRID: a general repository for interaction datasets. *Nucleic Acids Res*. 2006;34(suppl_1):D535–D539. doi:10.1093/nar/gkj109

16. Petar Velickovic, Guillem Cucurull, Arantxa Casanova, Adriana Romero, Pietro Lio, and Yoshua Bengio. Graph attention networks. In *International Conference on Learning Representations (ICLR)*, 2018.

17. J. Gilmer, S. S. Schoenholz, P. F. Riley, O. Vinyals, and G. E. Dahl, "Neural message passing for quantum chemistry," *arXiv preprint arXiv:1704.01212*, 2017.

18. Lamb J, Crawford ED, Peck D, et al. The connectivity map: Using gene-expression signatures to connect small molecules, genes, and disease. *Science (80- )*. 2006;313(5795):1929-1935. doi:10.1126/science.1132939

19. Muhan Zhang, Zhicheng Cui, Marion Neumann, and Yixin Chen. An end-to-end deep learning architecture for graph classification. In Proceedings of the AAAI Conference on Artificial Intelligence, volume 32, 2018.

20 Rex Ying, Jiaxuan You, Christopher Morris, Xiang Ren, William L Hamilton, and Jure Leskovec. Hierarchical graph representation learning with differentiable pooling. arXiv preprint arXiv:1806.08804, 2018.

21. Saurabh Verma and Zhi-Li Zhang. Graph capsule convolutional neural networks. arXiv preprint arXiv:1805.08090, 2018.

22. Discriminative embeddings of latent variable models for structured data. In International conference on machine learning, pages 2702–2711. PMLR, 2016.

23. Learning convolutional neural networks for graphs. In International conference on machine learning, pages 2014–2023. PMLR, 2016.